\documentclass[a4paper]{llncs}
\usepackage{times}
\usepackage{epsfig}
\usepackage{xspace}
\usepackage{subfigure}

\newlength{\dummy}
\newlength{\dumn}
\newlength{\dumnn}
\newlength{\dumx}

\newcommand{\nop}[1]{}

\newcommand{\Or}{\ensuremath{\vee}}
\newcommand{\derives}{\ensuremath{\ \mathtt{:\!-\ }}}

\newcommand{\dlv}{\textbf{\small{DLV}}\xspace}

\title{Toward the Implementation of Functions in the DLV System\\
(Preliminary Technical Report) }

\author{ {Francesco Calimeri\inst{}}\and {Nicola Leone\inst{}} \and{others}}\institute{Department of
Mathematics, University of Calabria}

\begin{document}

\maketitle

\begin{abstract}\label{abstract}
  This document describes the functions as they are treated in the
  \dlv system. We give first the language, then specify the main
  implementation issues.

\end{abstract}

\section{Language Specifications}\label{section:languageSpec}
In this section we briefly provide a formal definition of the
language.

\subsection{Syntax}\label{section:languageSpec.syntax}
A rule is in the form\\

{\em head} \derives {\em body}. \\

\noindent{where}

\begin{itemize}
\item[] {\em head} is a disjunction of {\em atoms};
\item[] {\em body} is a conjunction of {\em literals}.
\end{itemize}

\noindent{So we have}\\

{\em head}\ ::=\ {\em atom}\ [\ ``\Or''\ {\em head}\ ]

{\em body}\ ::=\ {\em literal}\ [,\ {\em body}\ ]

{\em literal}\ ::=\ [\ {\bf not}\ ]\ {\em atom}

{\em atom}\ ::=\ {\em predicate\_symbol}\ $\|$\ {\em predicate\_symbol}\ ``(''\ {\em terms}\ ``)''

{\em terms}\ ::=\ {\em term}\ [,\ {\em terms}\ ] \\

\noindent{and finally} \\

{\em term}\ ::=\ {\em constant}\ $\|$\ {\em variable}\ $\|$ $f$ ``('' $t_1$ ... $t_n$ ``)'' \\

\noindent{where}\ {\em $t_1$ ... $t_n$}\ \ are terms and\ \ {\em f}\ \ is a function symbol of arity\ \ {\em n}.

\section{Implementation Issues}\label{section:implementationIssues}
Functions implementation can be done involving only the grounding
module. Basically, we rewrite the rules containing some function
symbol in a specific manner, and set an order in their body; then we
properly control the match during the grounding. The resulting ground
program is passed to the Model Generator module, that does not need to
be changed.

\subsection{Rewriting Rules}\label{implementationIssues.rewritingRules}
All rules that do not contain functions are treated as usual, while
others are rewritten as follows. {\em Functions} are ``{\em
  flattened}'' in special body predicates, let's call them ``{\em
  function predicates}'', that are {\em built-in} predicates.  They
have the same arity as the original functions, plus one. All terms are
the same as those in them, but one: the first argument (``{\em
  id-argument}'') is an identifier representing the new term; it is
special, and it allows to ``link'' the new predicate symbol to the
ones that contain the original function.

\begin{example}\label{example:rewritingRules}
Let's consider the rule \\

$r$\ :\ \  p(s(X)) \derives\ \ a(X, $f$(Y,Z)). \\

\noindent{It} contains two function symbols,\ \ {\em s}\ \ and\ \ {\em f}. We create
two new predicates,\ {\em $s^*$}\ and {\em $f^*$}, and so rewrite the rule
as \\

$r'$\ :\ \  p(S) \derives\ \ a(X, F),\ \ $f^*$(F, Y, Z),\ \ $s^*$(S, X). \\

\noindent{The}
last arguments in the new predicates are the same as in the original
functions; the first one (in this case\ \ {\em S}\ and\ {\em F}\ \ for
$s^*$ and $f^*$ respectively)\ is their ``{\em id}'', and it appear
where necessary instead of the old function symbol (in this case in
the {\em p}\ and the {\em a}\ atoms).
\end{example}

\subsubsection{Optimization Note}\label{implementationIssues.reorderingBodies.optimizationNote}
If the same function appears in more than an atom, we should create
only one new function symbol. But it have to appear once or more than
once depending on the arguments.

\begin{example}\label{example:functionMultipleOccurrences}
The rule \\

$r$\ :\ \  p(s(X)) \derives\ \ q(s(X), Y). \\

\noindent{will} become \\

$r$\ :\ \  p(S) \derives\ \ q(S, Y),\ \ $s^*$(S, X). \\

\noindent{But} \\

$r$\ :\ \  p(s(X)) \derives\ \ q(s(Y), X). \\

\noindent{will} become \\

$r$\ :\ \  p(S) \derives\ \ q(S, X),\ \ $s^*$(S, X), $s^*$(S, Y).
\end{example}

\subsubsection{Implementation Note}\label{implementationIssues.rewritingRules.implementationNote}
The parser module should continue doing only its standard job, so it
should have only to recognize the {\em function predicates} as new
kind of tokens; in particular, as arguments of some predicate. Simply,
we have to remember that they are functions and lately rewrite them:
the job will be done by the rewriting module.

\subsubsection{Some Cares}\label{implementationIssues.rewritingRules.someCares}

\begin{itemize}
\item[A.]({\em Negations}) No particular attentions should be paid to
  the rules containing negations. The translation described above
  works properly.

\begin{example}\label{example:negation}
The rule \\

a(X)\ \derives\ p(X),\ {\bf not} ab(s(X)). \\

\noindent{is} rewritten as \\

a(X)\ \derives\ p(X),\ $s^*$(S, X),\ {\bf not} ab(S).
\end{example}

\item[B.]({\em Aggregates}) The ``flattening'' of rules containing
  aggregates, on the other hand, requires some care. It can be done in
  two ways:

\begin{itemize}
\item[1.] it is applied {\em after} the rewriting for the aggregates;
\item[2.] it is applied at the beginning; but, in this case, it has to
  be applied to the conjunctions inside the aggregates.
\end{itemize}

\noindent{We} think that the way (1.) is easier and more clean.

\begin{example}\label{example:aggregates}
  Consider the rule \\

$r$\ :\ \  a(X)\ \derives\ X = \#count( Y: p(s(Y)), q(Y) ). \\

\noindent{It} can be flattened choosing one of the following.
\begin{itemize}
\item[1.] Wait that is rewritten (during the parsing) to \\

$r_1$\ :\ \  a(X)\ \derives\ X = \#count( Y: aux(Y) ). \\
$r_2$\ :\ \  aux(Y)\ \derives\ p(s(Y)),\ q(Y). \\

Then flatten only $r_2$ as \\

$r'_2$\ :\ \ aux(Y)\ \derives\ p(S),\ q(Y),\ $s^*$(S, Y). \\
\item[2.] Flatten $r$ as \\

$r'$\ :\ \ a(X)\ \derives\ X = \#count( Y: p(S),\ q(Y),\ $s^*$(S, Y) ). \\

Then let it be rewritten to \\

$r'_1$\ :\ \  a(X)\ \derives\ X = \#count( Y: aux(Y) ). \\
$r'_2$\ :\ \ aux(Y)\ \derives\ p(S),\ q(Y),\ $s^*$(S, Y). \\
\end{itemize}
\end{example}

\item[C.]({\em Dependency Graph}) Function predicates are neglected in
  the dependency graph, in the same way it is already done with
  built-in predicates: they do not generate any arc in the $DG$.
\end{itemize}

\subsection{Reordering Bodies}\label{implementationIssues.reorderingBodies}
In the body-reordering, a function predicate should be inserted in the
first position such that:

\begin{itemize}
\item[a - ] either the {\em id} argument is bound, {\bf or}
\item[b - ] all previous arguments are bound, excepting at most the
  {\em id}.
\end{itemize}

\begin{example}\label{example:reorderingBodies_01}
  Let's consider the predicate\ \ $f^*$(Y, Z, F)\ \ in the rule $r'$
  of example~\ref{example:rewritingRules}. Here $f^*$ and $s^*$ should
  be put so that either\ $F$\ ($S$, respectively)\ is bound, or both\
  $Y$\ and\ $Z$\ ($X$, respectively)\ are bound: \\

  $r'$\ :\ \  p(S) \derives\ \ a(X, F),\ $f^*$(F, Y, Z),\ \ $s^*$(S, X). \\

  \noindent{Note} as the atom $a(X, F)$ is considered as first. So if the match
  does not fail (in the case it does $f^*$ and $s^*$ are not
  considered at all), $X$ and $F$ become both bound (or they already
  are, as constants).  So for $f^*$ we have the ``id-argument'' ($F$,
  in this case) bound; and for $s^*$ we have all arguments different
  from the {\em id} ($X$, in this case) bound.
\end{example}

\begin{example}\label{example:reorderingBodies_02}
  The rewritten rule \\

  $r'_1$\ :\ \  m(X, Y) \derives\ $s^*$(S, X, Y),\ k(S, T),\ p(W, Z, T). \\

  \noindent{has} the body not properly reordered: $s^*$ is considered as first,
  when neither the {\em id} argument nor all others are bound. A
  correct rewriting is the following: \\

  $r'_1$\ :\ \ m(X, Y) \derives\ k(S, T),\ $s^*$(S, X, Y),\ p(W, Z, T). \\

  \noindent{Here} the function predicate $s^*$ is in the first position such that
  (at least) one of the two conditions explained above is satisfied:
  in this case the ``id-argument" $S$ is always bound, if the matching
  process reaches $s^*$.
\end{example}

Actually, even if we place all function predicates at the end of each
body, everything should work; but we risk to loose performance
advantages due to the way the algorithm cut the search space of ground
rules. So we should put any function predicate in the {\em place} such
that at least one of the two previously exposed conditions holds.

To obtain what described, we can let the old algorithm work as usual,
but adding further checks to properly place function predicates.

We can put somewhere the ``functional'' atoms, for instance at the
very end of the body. Then we let the existing algorithm work as usual
to place one ``standard'' atom per time. Between two standard atoms'
placements, as soon as one took its place, we check, for each unplaced
functional atom, whether (at least) one of the two conditions holds;
if so, we put it after the standard atom just placed. It should be not
too difficult to check per each argument whether it occurs in at least
a placed atom or not. At the end, we are sure that all atoms were
properly placed.

\begin{example}\label{ex:reorderingBodies_03}
Let's examine the rule \\

$r$\ :\ \  p(X) \derives\ q(S, X),\ \ t(Y),\ \ $s^*$(S, Y). \\

\noindent{We} put initially $s^*$ at the end. The existing algorithm
chooses the first atom, say it $q$. Then we check $s^*$: it contains
$Y$ and its {\em id} $S$. Because the {\em id} appears in an already
placed atom, we can place also $s^*$ in the current position, removing
it from the last place. Then $t$ is placed, too, and we stop. Of
course, if the first atom placed was $t$, we could have placed $s^*$
anyway, beacuse all its arguments but the {\em id} were ok.

Basically it is possible to use the same trick we did for the
built-ins: we extend the heuristic function computing the ``weight''
of each atom to deal with ``function atoms''.  If one of those
conditions is verified, then the function atom has weight {\em
  -maxint}; otherwise its weight is {\em +maxint}.
\end{example}

\subsection{Grounding}\label{implementationIssues.grounding}
As already said, we would like to make functions transparent as much
as possible. So everything could stay the same (actually function
predicates can be viewed more or less like standard predicates),
excepting for a case during the matching algorithm.

With function predicates, it is possible to add ``tuples'' to some
table even if its symbol appears in the body, but not also in the head
(in fact, because of our ``flattening'', function predicates can not
appear in the head of a rule). So their ``tables'' can be filled only
by the facts (example: $f(s(1))$) or by some body occurrences. For
instance, if all arguments but the {\em id} are bound as in $f^*$(F,
a, b), then the match function should insert a new tuple (say
<a,b,$@_2$>) in $f^*$, with a new identifier ($@_2$, in this case) to
which the {\em id} should be bound.

This possibility is due to the fact that after flattening no function
predicates appear in the heads; but maybe they do in the original
rules. In these cases, we {\em have} to derive anyway.

\begin{example}\label{ex:grounding_01}
Let's have a look at the rule \\

$r$\ :\ \  p(s(X)) \derives\ q(X). \\

\noindent{Here} if $X$ matches with some value in the table of $q$,
we should derive what is in the head. But the flattened rule appear as

$r'$\ :\ \  p(S) \derives\ q(X),\ \ $s^*$(S, X). \\

\noindent{So} if we match $X$ for some value, if we do not find an
appropriate entry in $s^*$ we have to create it and succeed. Of course
{\em nothing} should be done if a matching tuple is present in $s^*$.
\end{example}

But if the function symbol did not appear in the head of the original
rule, we {\em should not} add new entries.

\begin{example}\label{ex:grounding_02}
With the rule \\

$r$\ :\ \  p(X) \derives\ q(X, s(Y)),\ \ t(Y). \\

if no matches are found for $q$, we should fail anyway. The rewritten
rule should appear as \\

$r'$\ :\ \  p(X) \derives\ t(Y),\ \ $s^*$(S, Y), q(X, S). \\

So if we have $Y$ bound, say $Y=b$, and in $s^*$ no tuples of kind
<b,\_> are found, we have to fail and come back to the choice of $Y$.
\end{example}

In practice, if the function predicate is not an ``head-predicate''
(in the sense just discussed), it has to be treated exactely as a
``standard'' predicate. No changes at all. If it is an
``head-predicate'', on the other hand, then 3 cases are possible, when
the matching function reaches it (these cases are always the same,
even if it is not an head predicate, but then nothing changes anyway):

\begin{itemize}
\item[1 - ] it has {\em id} argument bound, and not all the others are
  so;
\item[2 - ] it has all arguments bound, including the {\em id};
\item[3 - ] it has all arguments bound but the {\em id}.
\end{itemize}

It is easy to see as first two cases are the same, in our situation.
The function predicate has to be treated exactely like a ``standard''
one. Otherwise, in case $1$, we have to add a new tuple having the
values fixed to the bound values, and a {\em new} identifier. Of
course, this is a {\em try}: only if we reach the last body atom with
success we can, let's say, ``commit'' the new values. It should work
even if the same function with the same arguments appears in more than
one predicate, for instance one in the head and one in the body. This
will show us also the way we have to behave with respect to the
backtracking.

\begin{example}\label{ex:grounding_03}
Suppose that the body reordering in tail to the flattening gives for
the rule \\

$r$\ :\ \  p(s(X)) \derives\ t(X),\ \ q(s(X), Y). \\

\noindent{the} one

$r'$\ :\ \  p(S) \derives\ t(X),\ \ $s^*$(S, X), q(S, Y). \\

\noindent{Here} if we reach $s^*$, $X$ is bound (say to X=b) and $S$
is not; if no tuples of kind <b,\_> are found, we try to prepare a new
tuple <b, $next\_valid\_value\_for\_the\_identifier\_of\_s^*$>,
binding $S$ and passing to $q$. We will fail for sure because no such
a value for $S$ is in $q$ (it is new!!!). So we should backtrack, but
it is useless to try again with $s^*$: we have to jump it and come
back to $t$. Note as this seems to be consistent: we should have the
same result if the rule was ordered as

$r''$\ :\ \ p(S) \derives\ q(S, Y),\ \ $s^*$(S, X),\ \ t(Y). \\

\noindent{In} this case we succeed only if they match with an existing
value in $q$.
\end{example}

Since a ``function atom'' always has a key argument which is bound at
the time when it is instantiated (either the {\em id} or all others),
then the functional dependency guarantees that it cannot ``produce''
more than one match. Therefore, during the backward phase (i.e., if if
the match of a subsequent body atom failed) we should not retry to
match a function atom; but we should skip it and go to the previous
body atom.

Resuming: no changes if the predicate is not an ``head-predicate'';
pay attention only if it is {\em and} the {\em id }is not bound. About
backtracking, the function predicates have to be jumped. It is worth
to remember that even in the previous case ($3$), when the match
succeed because of an existing tuple, everything remain the same (no
new tuple creation), excepting the backtracking policy (do
backjumping). But it is necessary to provide an efficient way to know
whether a function atom comes from the head or from the body.

All these stuffs make sense with respect to the way to manage facts
containing functions. An example will clarify.

\begin{example}\label{ex:grounding_04}
  Consider\ $p(s(1)).$ Writing a sort of ``temporary rule'', only for
  meaning, we could see\ \ $p(S)$ :- $s^*(1, S)$.\ We should
  ``ground'' somehow that $S$. Following a kind of ``high-level''
  semantics, we should add a value in the table of $p$, and a value in
  the table of $s^*$. We could look-up the table for $s^*$: if a tuple
  of kind <1,\_> already there exists (say <1,$@_1$>) we add <$@_1$>
  to $p$ and do nothing else (think about a previous fact like
  $q(s(1)).$).  If not, we create a new identifier (say $@_3$), put it
  in $p$, and in addition we add the tuple <1,$@_3$> to $s^*$.
\end{example}

This choice will grant a single identifier for a single value of the
function. It should work even in case of arity $>1$.

\subsubsection{Implementation Note}\label{implementationIssues.grounding.implementationNote}
With respect to the choice about the parser's role, we can shortly
describe the behaviour of the rewriting module. It can add the tuples
for facts containing functions, while should simply rewrite the rules
in which some functions appear to be processed lately by the grounding
module.

\subsubsection{Odd Stuffs}\label{implementationIssues.grounding.oddStuffs}
Some considerations. \\

A function predicate born from a ``function atom'' appearing both in
the head predicate and in the body, has to be considered as a full
``body-predicate'': it can not produce new values. \\

Since the matching of a function predicate ``coming from head'' always
succeed (as exposed above), then we could avoid to match it during the
instantiation of the body (it seems useless); rather we match it
adding the corrisponding value only when we ``commit'' the grounding
instantiation of the rule. A possibility is to put those kind of
predicates at the very end of the body. \\

Once we've inserted the new values in the function tables, they are
true. When the program is simplified, removing all true atom from the
bodies after the grounding, the funcion atoms should be deleted. If we
reache the aim to treat the function predicates almost as standard
ones, we should not have to do a thing, because of the existing
simplifying methods. \\

Processing nested functions is almost transparent, and can be done
inside-to-outside or outside-to-inside. For instance, $p(s(f(1, a)),
2)$ can be first rewritten as\ \ $p(s(F), 2) \derives f^*(F, 1, a).$\ \
and then as\ \ $p(S, 2) \derives f^*(F, 1, a),\ s^*(F, S).$ \\

About the problem of no-termination due to the nesting, we can give
the user the possibility to guarantee the termination. This can be
done by telling the system when it should stop the generation of
function symbols. That is, the user can tell: please dont generate
functions having nesting level greater than $k$. The nesting level is
the maximum number of nested ``subterms'' which are in a term. This
mean $0$ for a constant, and so on: for instance $1$ for s(a,b), $3$
for f(s(t,w(a)), f(b,c)), etc. The option will be something like
$-maxNesting=k$. In order to be able to support this option, we need
to store the nesting level of each term. At the beginning, we set it
to $0$ for the constants. Whenever we create a new function {\em id}
(i.e., we insert a new element in a function table), we look at its
arguments, we check what is the maximum nesting level ``l'' among
them, and we set to ``l+1'' the nesting level of the new {\em id}.  If
$maxNesting$ has been set to $k$, and the new term has nesting $>k$,
then the insertion fails. This means that even the set predicates
coming from the flattening of the head terms may fail. If the option
is not set, no checks are performed, so the termination is not
guarantee and it is under the responsability of the programmer.

This leads us to process the functions inside-to-outside.  For
instance\ \ $f(q(X, (w(Y), g(Y)))$\ \ gives first $g$ and $w$ at level
$1$, then $q$ at level $2$, and at last $f$ at level $3$ (supposing
that Y is bound to a constant).


\end{document}